\documentclass[conference]{IEEEtran}
\IEEEoverridecommandlockouts
\usepackage{cite}
\usepackage{amsmath,amssymb,amsfonts}
\usepackage{algorithmic}
\usepackage{graphicx}
\usepackage{textcomp}
\usepackage{xcolor}
\usepackage{booktabs}
\usepackage{multirow}
\usepackage{tabularx}
\usepackage{multicol}
\usepackage{orcidlink}
\usepackage{tikz}
\usepackage{float}
\usepackage{adjustbox}
\usepackage{hyperref}
\usepackage{colortbl}  

\def\BibTeX{{\rm B\kern-.05em{\sc i\kern-.025em b}\kern-.08em
    T\kern-.1667em\lower.7ex\hbox{E}\kern-.125emX}}

\begin{document}

\title{A Pluggable Multi-Task Learning Framework for Sentiment-Aware Financial Relation Extraction\\

\author{\IEEEauthorblockN{Jinming Luo}
\IEEEauthorblockA{
\textit{Southwestern University of Finance and Economics}\\
Chengdu, China \\
42253017@smail.swufe.edu.cn}
\and
\IEEEauthorblockN{Hailin Wang*}
\IEEEauthorblockA{
\textit{Southwestern University of Finance and Economics}\\
Chengdu, China \\
\textit{Kash Institute of Electronics and Information Industry}\\
{Kashi, Xinjiang, China}\\
wanghl@swufe.edu.cn}}

\thanks{*Corresponding author. We thank anonymous reviewers for their insightful feedback that help improve the paper. This research work is supported by the Natural Science Foundation of Xinjiang Uyghur Autonomous Region, China (No. 2023D01B12, 2024D01B07), the Sichuan Province Science and Technology Support Program, China (No. 2023NSFSC1412), the National Natural Science Foundation of China (No. 62376228), the Student Research Initiative Program of SWUFE-UD Institute.}
}
\maketitle

\begin{abstract}

Relation Extraction (RE) aims to extract semantic relationships in texts from given entity pairs, and has achieved significant improvements. However, in different domains, the RE task can be influenced by various factors. For example, in the financial domain, sentiment can affect RE results, yet this factor has been overlooked by modern RE models. To address this gap, this paper proposes a Sentiment-aware-SDP-Enhanced-Module (SSDP-SEM), a multi-task learning approach for enhancing financial RE. Specifically, SSDP-SEM integrates the RE models with a pluggable auxiliary sentiment perception (ASP) task, enabling the RE models to concurrently navigate their attention weights with the text's sentiment. We first generate detailed sentiment tokens through a sentiment model and insert these tokens into an instance. Then, the ASP task focuses on capturing nuanced sentiment information through predicting the sentiment token positions, combining both sentiment insights and the Shortest Dependency Path (SDP) of syntactic information. Moreover, this work employs a sentiment attention information bottleneck regularization method to regulate the reasoning process. Our experiment integrates this auxiliary task with several prevalent frameworks, and the results demonstrate that most previous models benefit from the auxiliary task, thereby achieving better results. These findings highlight the importance of effectively leveraging sentiment in the financial RE task.


\end{abstract}

\begin{IEEEkeywords}
Relation Extraction, Implicit Sentiment Label, Shortest Dependency Path, Attention Mechanism
\end{IEEEkeywords}

\section{Introduction}

Relation Extraction (RE) is a critical task in Natural Language Processing (NLP) that identifies semantic relationships between entities. It has broad applications across various domains, including question answering \cite{singhal2025toward}, knowledge graph construction \cite{2024Knowledge}, and information retrieval \cite{chen2024gap}. In the financial sector, the increasing influence of large language models \cite{wang2024causalbench} has made RE increasingly vital for financial analysis, as it enables the extraction of key relationships from financial texts. However, traditional RE methods \cite{soares2019matching} often struggle with domain-specific challenges, particularly in financial contexts.

Financial RE is essential for analyzing financial documents and economic trends, such as linking key performance indicators (KPIs) \cite{deusser2022kpi}, supporting financial decision-making \cite{yu2024fincon}, and analyzing Securities and Exchange Commission (SEC) reports \cite{kaur2023refind}. Market sentiment, which reflects investors' expectations and attitudes toward future developments, can significantly shape the interpretation of relationships between entities \cite{azimi2021positive}. For example, optimistic financial statements may signal potential growth, attracting investment, while pessimistic tones can indicate risks, leading to a decline in stock prices.

Despite its importance, effectively incorporating sentiment into RE remains challenging due to the complexity of financial texts,which often contain intricate sentence structures \cite{vardhan2023imetre,kaur2023refind}, long-dependency relations \cite{pasch2023ahead}, and detailed economic facts. In this context, the Shortest Dependency Path (SDP) \cite{xu2015classifying} offers a valuable tool for semantic analysis, by focusing on the most direct relationships between entities. In addition to improving RE accuracy, SDP also captures sentiment, as it filters out irrelevant words and highlights core linguistic structures connecting entities. In investment-related texts, market reactions, and corporate earnings forecasts, SDP can link specific entities and actions to emotional cues, offering a structured understanding of sentiment-sensitive relationships.

\begin{figure}[H]
\centering
\includegraphics[scale=0.3]{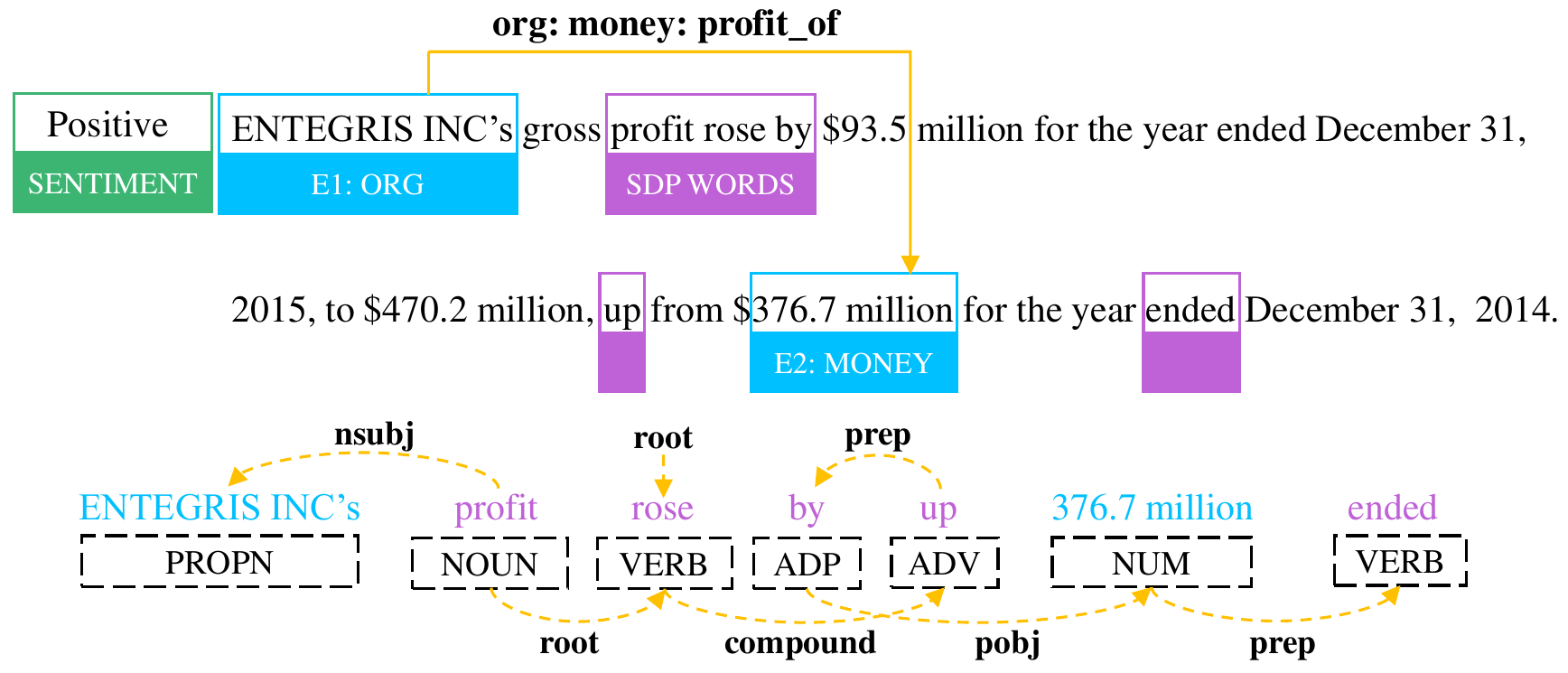}
\vspace{-0.6cm}
\caption{Example sentence from the REFinD dev dataset. The entities ``ENTEGRIS INC’s" (organization) and ``\$376.7 million" (money) are linked by the “profit\_of” relation, which is strongly associated with positive sentiment.}
\label{figure0}
\end{figure}

Consider the financial statement shown in Figure \ref{figure0}. SDP isolates key terms such as ``ENTEGRIS INC's", ``profit", ``rose", and ``up from", which directly connect the company to its financial features. ``Profit" serves as the subject noun (\textit{nsubj}), marking ``ENTEGRIS INC's" as the entity undergoing the change. The verb ``rose" labeled as the root, centrally defines the action of increase, indicating positive financial growth. The adverb ``up" functioning as a phrasal verb particle (\textit{compound:prt}), emphasizes the direction of the increase. This SDP analysis highlights the grammatical roles that convey the company's fiscal improvement, thereby capturing its success within the specified period. Sentiment analysis identifies the emotional tone as ``positive", indicating favorable financial growth, which can influence stakeholders' perceptions and market reactions. This precise parsing, coupled with the detected sentiment, provides a deeper layer of context that enriches the extraction of the financial relationship ``org: money: profit\_of".

Building on these insights, this paper introduces the Sentiment-aware-SDP-Enhanced Module (SSDP-SEM). SSDP-SEM innovatively integrates a multi-task learning architecture with specialized modules designed to manage Implicit Sentiment Label (ISL) and performs a pluggable auxiliary sentiment perception (ASP) task with RE task. To incorporate sentiment, SSDP-SEM concatenates the sentiment token of the text with itself, providing a high-level tone indicator for a context-aware representation. To simultaneously capture key semantic actions, the module extracts SDP tokens, effectively reducing noise and isolating critical relationships for high-precision financial text processing. These sentiment and SDP tokens together form the ISL, supervising the ASP task through fine-tuning various attention mechanisms, allowing the model to focus on the most pertinent part of the text. Additionally, SSDP-SEM employs a sentiment attention information bottleneck (SAIB) regularization method within its final layers. This method uses a sentiment attention mechanism to adaptively select key features while disregarding redundancy, further enhancing the model's focus. 


The main contributions of this paper are summarized as follows:
\textbf{1)} The multi-task learning architecture uniquely combines sentiment analysis with SDP, enhancing its ability to interpret both the sentiment dimensions and structures of complex financial texts.
\textbf{2)} The work devises novel supervisory signals ISL to refine the attention mechanism of various models, enabling precise focus on the most relevant syntactic paths and sentiment cues.
\textbf{3)} The module introduces a sentiment attention information bottleneck regularization method within the final layers, allowing the model to adaptively select key features while ignoring redundancy.

\section{Related Work}
RE in NLP has made significant strides through innovative methods and models. Recent advances in RE focus on enhancing semantic representation, incorporating auxiliary linguistic features and multi-task learning, and developing domain-specific solutions for financial text analysis  \cite{sun2022lexicalized,sinha2022sentfin,montariol2022multi}. Our work intersects these research streams by integrating syntactic dependency analysis with sentiment-aware learning \cite{ye2022sentiment}, addressing unique challenges in financial RE.

One such advancement is the introduction of the Entities and Mentions Gradual Enhancement (EMGE) framework, which integrates contextual and structural information to enhance entity representations and capture long-distance dependencies in document-level extraction tasks \cite{chen2025emge}. Additionally, incorporating Named Entity Recognition (NER) and Part-Of-Speech (POS) tagging into pre-trained language models has proven effective, particularly in improving performance on financial RE datasets \cite{li2024enhancing}. Furthermore, Lexicalized Dependency Paths (LDPs), a computationally efficient rule-based method, utilize dependency paths with lexical and syntactic information to reduce reliance on large-scale training data for RE \cite{sun2022lexicalized}.

Alongside these advancements, sentiment analysis has emerged as a critical enhancer for financial RE, where affective cues fundamentally shape entity relationships \cite{mondal2018relation}. Financial texts inherently contain sentiment-laden indicators—such as profit anticipation, risk perception, and market confidence—that directly influence relational semantics \cite{sinha2022sentfin}. Recent progress shows that multitask architectures that jointly optimize sentiment prediction and relation classification achieve superior performance in financial domains \cite{montariol2022multi}. Building on this, transformer-based models now strategically balance syntactic patterns with emotional signals, using dependency structures to align sentiment valence with relational contexts \cite{cheruku2024sentiment}.

The importance of SDP in enhancing RE is also gaining increasing recognition \cite{wang2024enhancing}. For instance, a RoBERTa-based approach for financial RE was proposed, which incorporates SDP features to capture syntactic dependencies and improve the accuracy by modeling the intricate connections between entities \cite{qiu2023simple}. Studies have further shown that leveraging SDP significantly enhances the extraction of key dependency relationships, improving the handling of complex sentence structures and specialized financial terminology \cite{rajpoot2023nearest}.

Despite these advances, current approaches exhibit two critical limitations: 1) existing sentiment-aware models inadequately leverage syntactic structure, and 2) SDP implementations neglect affective cue weighting. Our framework bridges this divide through dual-channel optimization of syntactic patterns and sentiment valence.


\section{Methodology}

\begin{figure}[h]
\centering
\includegraphics[scale=0.085]{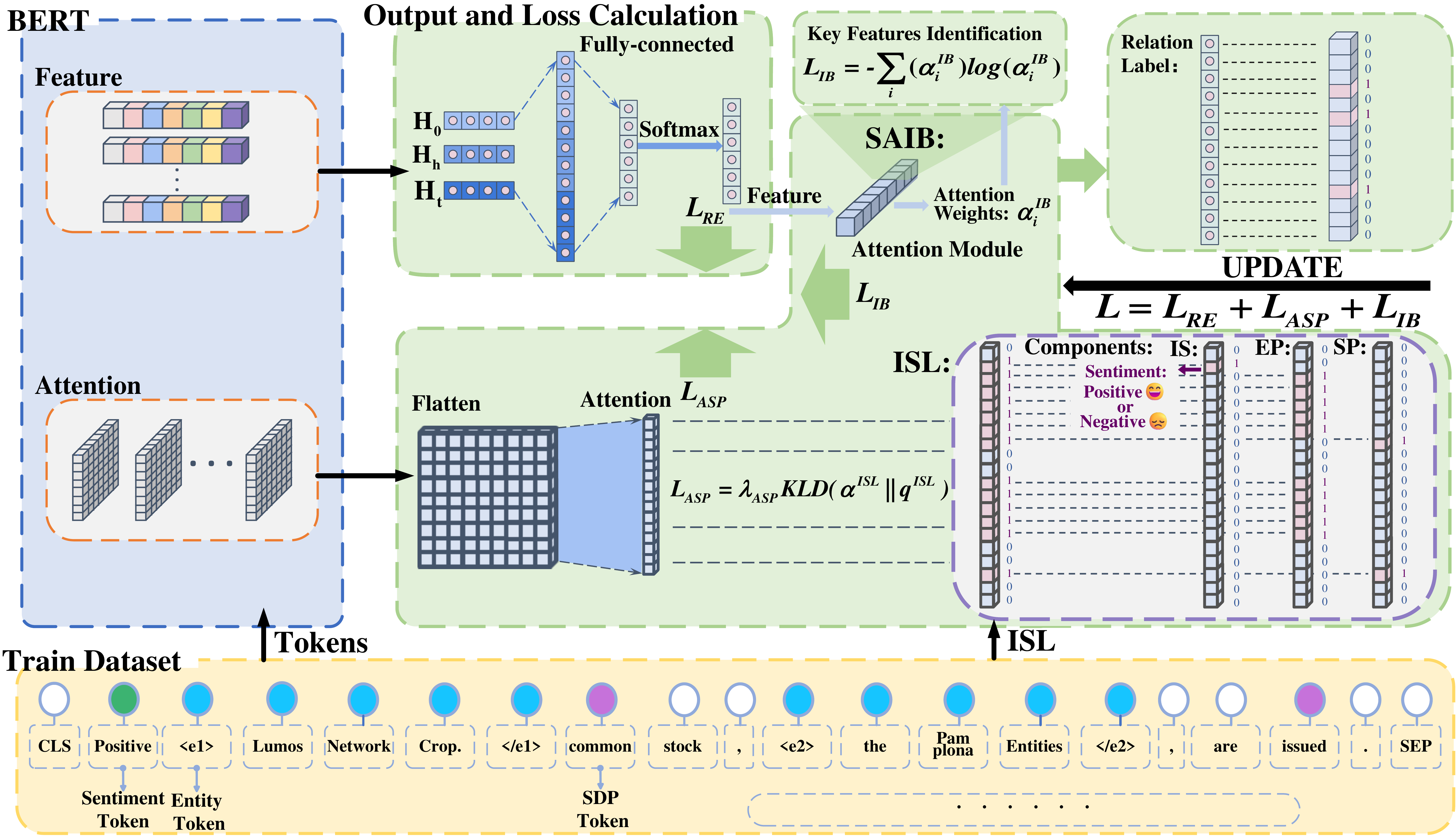}
\caption{Architecture of SSDP-SEM. This architecture presents a way of using the multi-task learning pluggable module.}
\vspace{-0.3cm}
\label{figure1}
\end{figure}

Our method focuses on optimizing financial RE by integrating ASP tasks to enhance the performance through sentiment analysis in a multi-task architecture. We exploit the ASP task to integrate with various attention-based RE models, sharing the same model parameters. The ASP predicts the positions of ISL tokens, while the original RE task continues to predict the relations between entity pairs, enhanced by our SAIB regularization loss. These two tasks are trained in parallel. Figure \ref{figure1} illustrates the entire module, which consists of three components:

(1) \textbf{ISL Constructor}: generates sentiment tokens and extracts corresponding SDP tokens for each sentence to form a unified ISL signal.
(2) \textbf{ASP with ISL Signal:} utilizes the ISL as supervisory signals and applies a merged attention weight as supervised features for the ASP task.
(3) \textbf{RE with SAIB Regularization:} applies a sentiment attention mechanism to encourage the model to focus more on a small number of key features and ignore the irrelevant redundant information.




\subsection{Task Definition}
\textbf{Financial RE}: Given a sentence \(S\) in financial texts consisting of words \( \mathcal{X} = \{x_1, x_2, ..., x_n\} \) and a pair of entities \( E_S = (e_s, e_o) \), the goal is to predict the relation \( r \in R \) between the two entities, where \( R \) is a predefined set of relations. In this work, we incorporate our module with various baseline models, thus the task definition is the same as these baselines.

\textbf{ASP task}: A sentiment token $\mathcal{X}_{sen}$ is inserted into each instance, and the SDP token sequence $\mathcal{X}_{sdp}$ is extracted from the original text $\mathcal{X}$. The sentiment token is then joined with the SDP tokens to form the ISL, which automatically annotates the token position indices in the instance. The ASP task trains the model to predict all positions of the ISL tokens. Through this task, all baselines can be trained to recognize the distribution of ISL token positions, and further learn the sentiment semantics.

\begin{figure*}[htp]
\centering
\includegraphics[scale=0.3]{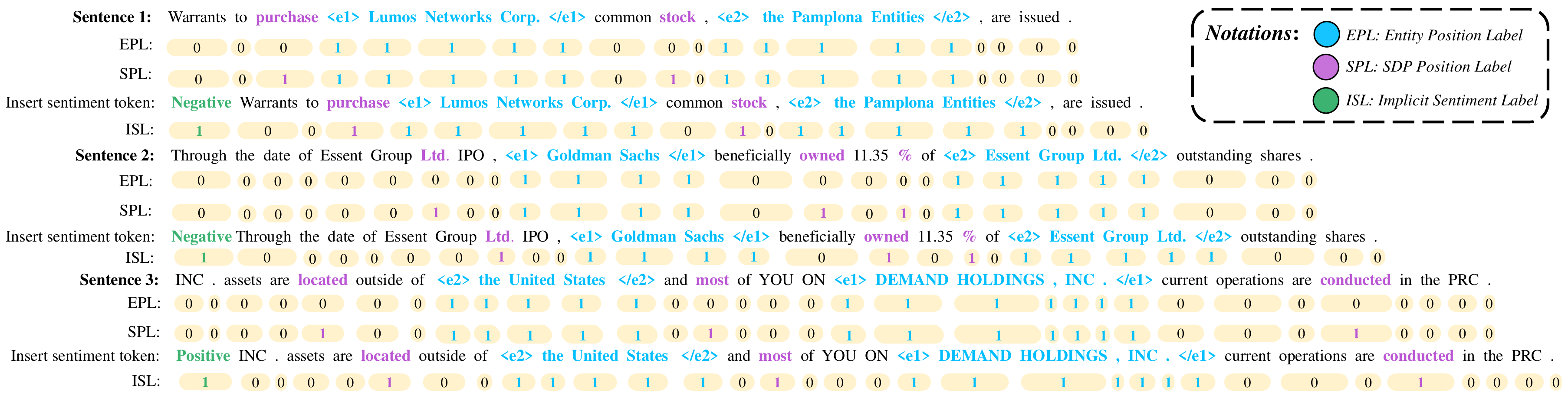}
\vspace{-0.6cm}
\caption{Examples of ISL and their corresponding EPL and SPL. ${\text{EPL}}$ focuses on entity token position, ${\text{SPL}}$ combines SDP token positions with EPL, and ${\text{ISL}}$ incorporates EPL and SPL.}
\label{figure2}
\end{figure*}

\subsection{ISL constructor}
\textbf{Sentiment Token}: As mentioned earlier, sentiment tokens significantly influence the analysis of text semantics, especially in financial documents. In our approach, sentiment tokens are implicitly defined as either ``positive" or ``negative". One of these emotional tokens is strategically inserted into the text to represent the overall sentiment of the financial statement. This insertion simplifies the integration of sentiment analysis. We first use \textit{senta\_lstm} \cite{ma2019paddlepaddle} to obtain the sentiment result \(\mathcal{X}_{\text{sen}} = \text{senta\_lstm}(\mathcal{X}) \). And then insert this token into the original text as \( \mathcal{X} = (\mathcal{X}_{\text{sen}}: \mathcal{X}) \).


\textbf{SDP Tokens}: To acquire the SDP token positions distribution, we utilize Spacy \footnote{https://spacy.io/} to obtain the syntactic dependency tree, and further extract the SDP with given entity pair as $\mathcal{X}_{sdp}=\{e_s,x_i \dots, root,\dots, x_j, e_o \}$. SDP tokens reduce the complexity of the parsing process by focusing only on the most relevant parts of the sentence structure, and these tokens also contribute to the understanding of text sentiment.

\textbf{ISL}: To integrate sentiment and SDP information into the model, we construct the ISL signal, denoted as \(Q^{ISL}\)\, which identifies the positions of tokens. It is defined based on the following criteria: 
\begin{equation}
    {Q^{ISL}_{i}} = 
    \begin{cases} 
        1 & \text{if } x_i \in \mathcal{X}_{sdp} \cup \mathcal{X}_{sen} \\
        0 & \text{otherwise}
    \end{cases}
    \label{eq:isl_label}
\end{equation}
To calculate the supervisory signal, we normalize the binary vectors \(Q^{ISL}\)\  as follows:
\begin{equation}
    {q^{ISL}} = \frac{{{Q^{ISL}}}}{{\sum\nolimits_{i = 1}^{\left| \mathcal{X} \right|} {{Q^{ISL}}_{i}}} }
\end{equation}
where $\left| \mathcal{X} \right|$ is the length of the sentence \(S\). This normalization ensures that the attention mechanism focuses proportionally on the important tokens, based on both the syntactic and emotional relevance.

In addition, for comparing this \textbf{ISL}, we also devise another two labels: Entity Position Label (\textbf{EPL}) and SDP Position Label (\textbf{SPL}), use Entity Position (EP) and SDP Position (SP) as indicators of these labels, as defined in Formula \ref{eq:isl_label}. Our ISL integrates these two positions along with Implicit Sentiment (IS) token position. Three examples of these labels are presented in Figure \ref{figure2}.


\subsection{ASP with ISL Signal}
ISL token positions encompass rich semantics for financial RE, particularly for sentiment tokens paired with SDP tokens, which provide vital clues about the sentiment of the text. The goal of the ASP task is to guide the RE model to focus more attention on these ISL tokens. Thus, we define the ISL as the supervisory signal for ASP, expecting the ASP task to learn the distribution of ISL token positions and predict all ISL token positions. Furthermore, ASP supervises the attention weights $A_{ISL}$ from the pre-trained language model as the predicted features.

\textbf{Supervised Attention for ASP}: In our method, $A_{ISL}$ is derived from the multi-head attention, which is commonly employed in transformer-based models such as BERT and GPT, enabling the model to capture diverse aspects of the relationships between tokens. The formula for multi-head attention is:
\begin{equation}
    \mathbf{\alpha}^{mha} = \text{softmax}\left(\frac{QK^T}{\sqrt{d_k}}\right)
\end{equation}
where $\mathbf{\alpha}^{mha}$ represents the attention weights, \(Q\) and \(K\) are the query and key matrices, respectively, with \(d_k\) being the dimension of the key vectors.

To construct the attention-based supervision, the attention weights are synchronized with the locations of ISL tokens. Initially, attention weights are derived from the model's final three layers, aggregating the attention values of each token across all attention heads. For each token, the average attention weight is computed as:
\begin{equation}
\alpha^{avg}_i = \frac{1}{H}\sum_{h=1}^H \alpha_i^{mha}
\end{equation}
where \(H\) is the number of attention heads. The supervised signal \(\alpha^{ISL}\) is then generated by multiplying these average attention weights with the ISL \(Q^{ISL}\) in Hadamard manner for each token, guiding it to concentrate on tokens deemed most pertinent based on their syntactic and emotional relevance.

\begin{equation}
\alpha^{ISL} = \alpha^{avg} \odot Q^{ISL}
\end{equation}

During the training process, this normalized vector \(q^{ISL}\) serves as a supervised signal to guide the model’s attention towards ISL tokens, thereby improving its predictive ability for implicit relationships and token positions.

\textbf{Loss for ASP}: To accommodate the new supervisory signals, this module requires a corresponding loss function. We introduce a Kullback-Leibler Divergence (KLD) loss to measure the discrepancy between the predicted distribution of token positions and the target distribution from ISL. The loss is calculated as:
\begin{equation}
    {L_{ASP}} = \lambda_{asp}KLD(\alpha^{ISL}\left\| q^{ISL} \right.)
\end{equation}
where $\alpha^{ISL} $ represents the attention-based supervised signal, ${q^{ISL}}$ indicates the distribution of ISL token positions, and $\lambda_{asp}$ is a hyperparameter that controls the strength of the attention to influence the results.




\subsection{RE with SAIB Regularization}
Introducing the ISL may disturb the original model's final classification result, as ASP is a pluggable auxiliary task. Therefore, we want a regularization to minimize the impact of auxiliary tasks and increase the influence of emotional factors on the expected experimental results, so we introduce the SAIB.


\textbf{SAIB}: By leveraging the attention mechanism on the final features, the model adaptively identifies key features and reduces redundancy, ensuring that only the most vital features are retained. This is achieved by applying an attention module that weights features and limits the entropy of the attention weights, promoting the model’s focus on a few essential features while minimizing noise. In other words, we want to make this attention have a sparse attention distribution. This process calculates the attention as follows:

\begin{equation}
    r_i = f(x_i, x_{sen}) = W_{\alpha}[{r}_{base}:{r}_{sen}]+b_{\alpha}
    \label{eq_ib}
\end{equation}
\begin{equation}
    \alpha_i^{IB} = \frac{exp(e_i)}{\sum_jexp(e_j)}
\end{equation}
where ${W}_{\alpha}$ is the learned weight matrix, and $[{r}_{base}:{r}_{sen}]$ represents the concatenation of the final feature(from the baseline in our work) with the sentiment token features.
The attention weights $\alpha_i^{IB}$ are calculated using the \textit{softmax} function applied to the attention scores $r_i$.

\textbf{Loss for SAIB}: To encourage a sparse attention distribution, we use entropy regularization, aiming to minimize the entropy of the attention weights. This forces the model to focus its attention on a small set of key features, resulting in a more concentrated attention distribution.
The sparsity is enforced by minimizing the entropy of the attention weights:
\begin{equation}
    L_{IB} = -\sum_i(\alpha_i^{IB})log(\alpha_i^{IB})
\end{equation}
This regularization function gives the final prediction feature penalty to those non-ISL token positions, which will be incorporated into our final loss function.


\subsection{Classifier and Final Loss}
\textbf{RE Classifier}: As mentioned, we use our module as a pluggable component to enhance the baseline models. Thus, we integrate the baseline output features with the previously introduced SAIB to obtain the final refined features, as follows:
\begin{equation}
    \mathbf{P}_i=\sigma(\sum_i\alpha_i^{IB}x_i)
\end{equation}
where $r_i$ is the output feature of the baselines, $\mathbf{P}_i$ is the probability of the i-th relation, $\sigma$ is the activation function of the corresponding baseline. 

\textbf{Final Loss Function}:
To optimize the model effectively, our final loss function combines the RE, ASP tasks with SAIB regularization into a unified global loss function, ensuring that the model learns from all implicit semantic information. Therefore, we design the final loss function as follows:
\begin{equation}
    L = L_{RE} +  L_{ASP} + L_{IB}
\end{equation}

This combined loss function enables the model to learn from both entity pair relations and the implicit semantic token position distribution, improving its overall performance on financial RE by effectively perceiving the sentiment of the text.





\section{Experiments}

\subsection{Experimental Setting}

\textbf{Dataset and Metric.} To investigate the influence of sentiment factors, this study evaluates the proposed method on the financial - related dataset REFinD (with 22 relations, Dev set of 4306, Train set of 20070, Test set of 4300) \cite{kaur2023refind} and the general - purpose dataset TACRED (with 42 relations, Dev set of 22631, Train set of 68124, Test set of 15509) \cite{alt2020tacred}. REFinD, sourced from the US SEC website, focuses on financial texts where sentiment significantly shapes entity relations like profit and loss. TACRED, primarily derived from news articles with formal language and complex sentence structures, suits approaches like SDP. These two datasets assess the effectiveness of the SSDP-SEM method, reporting accuracy, precision, recall, and F1 score as primary metrics.

\textbf{Baselines.} For REFinD, we included a variety of pretrained and domain-specific models—FinBERT \cite{liu2021finbert}, SpanBERT \cite{joshi2020spanbert}, DistilBERT \cite{sanh2019distilbert}, BERT \cite{devlin2018bert}, and RoBERTa-Large \cite{pasch2023ahead}—to capture financial text semantics, supplemented by syntax-aware methods such as SDP-DEP-NER \cite{rajpoot2023nearest} and TrNP \cite{li2024enhancing} for structural modeling. Additionally, we reproduced several attention-based architectures suited for our pluggable module (Att-Bi-LSTM \cite{zhou2016attention}, RE-Improved \cite{zhou2021improved}, R-BERT \cite{wu2019enriching}, and Casual \cite{wang2023causal}) to cover diverse approaches ranging from domain adaptation to syntactic parsing, thereby enabling a comprehensive evaluation on financial data. 

Similarly, for the auxiliary general-purpose TACRED dataset \cite{alt2020tacred}, we incorporated a position-aware model (PALSTM \cite{zhang2017position}), graph-based methods (CGCN \cite{zhang2018graph}, GCN \cite{guo2019attention}), and noise-robust approaches (PURE \cite{zhong2020frustratingly}, Clean-LaVe \cite{wang2024use}). We also tested the same reproduced attention-based architectures (Att-Bi-LSTM, RE-Improved, R-BERT, and Casual) on TACRED to verify whether the sentiment-aware enhancements observed in the financial domain could similarly benefit general RE tasks. Due to our focus on the influence of sentiment factors and limitations of some model architectures, we have not validated the effectiveness of the latest models.

\begin{table}[h]
\centering
\caption{Comparison of baseline models equipped with SSDP-SEM on the REFinD dataset. The symbol - indicates the metric was not provided in the original paper, \dag\ denotes the baseline models we reproduced, and +S\ddag\ indicates the models with SSDP-SEM.} \label{table2}
\vspace{5pt}
\vspace{-0.4cm}
\begin{tabular}{>{\raggedright\arraybackslash}p{2.45cm}   >{\centering\arraybackslash}p{1cm}  >{\centering\arraybackslash}p{1.15cm}  >{\centering\arraybackslash}p{1.15cm}  >{\centering\arraybackslash}p{1cm}}
\hline
\toprule
\multirow{2}{*}{Methods} & \multicolumn{4}{c}{REFinD} \\
\cmidrule{2-5}
 & Acc(\%) & Prec(\%) & Rec(\%) & F1(\%) \\
\midrule
\rowcolor[gray]{0.7} 
\textit{Baselines}& & & & \\ 
\midrule
BERT\cite{devlin2018bert} &-&-&-& 65.8 \\ [0.5ex]
DistilBERT \cite{sanh2019distilbert} &-&-&-& 63.6 \\ [0.5ex]
SpanBERT \cite{joshi2020spanbert} &-&-&-& 65.7 \\ [0.5ex]
FinBERT \cite{liu2021finbert} &-&-&-& 64.8 \\ [0.5ex]
RoBERTa-Large \cite{pasch2023ahead} &-&-&-& 72.6 \\ [0.5ex]
SDP-DEP-NER \cite{rajpoot2023nearest} &-&-&-& 76.3 \\ [0.5ex]
TrNP\cite{li2024enhancing} & - & - & - & 77.2 \\ [0.5ex]
\midrule
Att-Bi-LSTM\dag \cite{zhou2016attention} & 75.26 & 73.84 & 72.04 & 72.93 \\ [0.5ex]
R-BERT\dag \cite{wu2019enriching} & 77.28 & 74.63 & 74.60 & 74.61 \\ [0.5ex]
RE-Improved\dag \cite{zhou2021improved} & 77.10 & 75.11 & 74.21 & 74.66 \\ [0.5ex]
Casual\dag \cite{wang2023causal} & 79.48 & 80.27 & 72.29 & 76.07 \\ [0.5ex]
\midrule
 \rowcolor[gray]{0.7} 
\textit{Models+SSDP-SEM} & & & & \\ 
\midrule
Att-Bi-LSTM+S\ddag & 76.92\tiny{(\textbf{+1.66})} & 75.27\tiny{(\textbf{+1.43})} & 73.28\tiny{(\textbf{+1.24})} & 74.26\tiny{(\textbf{+1.33})} \\[0.5ex]
R-BERT+S\ddag & 78.54\tiny{(\textbf{+1.26})} & 76.66\tiny{(\textbf{+2.03})} & 74.64\tiny{(\textbf{+0.04})} & 75.64\tiny{(\textbf{+1.03})} \\[0.5ex]
RE-Improved+S\ddag & 77.61\tiny{(\textbf{+0.51})} & \textbf{77.87}\tiny{(\textbf{+2.76})} & 72.29\tiny{(-1.92)} & 74.98\tiny{(\textbf{+0.32})} \\[0.5ex]
Casual+S\ddag & \textbf{79.67}\tiny{(\textbf{+0.19})} & 78.65\tiny{(-1.62)} & \textbf{75.53}\tiny{(\textbf{+3.24})} & \textbf{77.06}\tiny{(\textbf{+0.99})} \\[0.5ex]
\hline
\toprule
\end{tabular}
\vspace{-0.6cm}
\end{table}

\begin{table}[h]
\centering
\caption{Comparison of baseline models equipped with SSDP-SEM on the TACRED dataset. The symbols - \dag \ddag are the same meaning as Table \ref{table2}} 
\label{table2.5}
\vspace{5pt}
\vspace{-0.4cm}
\begin{tabular}{>{\raggedright\arraybackslash}p{2.35cm}   >{\centering\arraybackslash}p{1.15cm}  >{\centering\arraybackslash}p{1.15cm}  >{\centering\arraybackslash}p{1.15cm}  >{\centering\arraybackslash}p{1cm}  }
\hline
\toprule
\multirow{2}{*}{Methods} & \multicolumn{4}{c}{TACRED} \\
\cmidrule{2-5}
 & Acc(\%) & Prec(\%) & Rec(\%) & F1(\%) \\
\midrule
\rowcolor[gray]{0.7} 
\textit{Baselines} & & & & \\ 
\midrule
PA-LSTM \cite{zhang2017position} &-& 67.8 & 64.6 & 66.2 \\ [0.5ex]
CGCN \cite{zhang2018graph} &-& 65.2 & 60.4 & 62.7 \\ [0.5ex]
GCN \cite{guo2019attention} &-& 69.2 & 60.3 & 64.5  \\ [0.5ex]
PURE \cite{zhong2020frustratingly} &-& - & - & 69.7 \\ [0.5ex]
Clean-LaVe \cite{wang2024use} &-&-&-& 63.4 \\[0.5ex]
\midrule
Att-Bi-LSTM\dag \cite{zhou2016attention} & 84.70 & 67.54 & 45.50 & 54.38 \\ [0.5ex]
R-BERT\dag \cite{wu2019enriching} & 87.68 & 69.47 & 66.53 & 67.97 \\ [0.5ex]
RE-Improved\dag \cite{zhou2021improved} & 87.89 & 70.88 & 64.14 & 67.89 \\ [0.5ex]
Casual\dag \cite{wang2023causal} & 87.93 & 71.32 & 65.29 & 68.17 \\ [0.5ex]
\midrule
\rowcolor[gray]{0.7} 
\textit{Models+SSDP-SEM} & & & & \\ 
\midrule
Att-Bi-LSTM+S\ddag & 84.64\tiny{(-0.06)} & 65.67\tiny{(-1.87)} & 47.76\tiny{(\textbf{+2.26})} & 55.30\tiny{(\textbf{+0.92})} \\[0.5ex]
R-BERT+S\ddag & \textbf{88.14}\tiny{(\textbf{+0.46})} & \textbf{71.60}\tiny{(\textbf{+2.13})} & 66.44\tiny{(-0.09)} & 68.92\tiny{(\textbf{+0.95})} \\[0.5ex]
RE-Improved+S\ddag & 87.82\tiny{(-0.07)} & 68.83\tiny{(-2.05)} & \textbf{68.41}\tiny{(\textbf{+4.27})} & 68.66\tiny{(\textbf{+0.77})} \\[0.5ex]
Casual+S\ddag & 88.00\tiny{(\textbf{+0.07})} & 71.50\tiny{(\textbf{+0.18})} & 67.05\tiny{(\textbf{+1.76})} & \textbf{69.19}\tiny{(\textbf{+1.02})} \\[0.5ex]
\hline
\toprule
\end{tabular}
\vspace{-0.8cm}
\end{table}

\subsection{Experimental Results}

Tables \ref{table2} and \ref{table2.5} show the comparison result of our method (SSDP-SEM) with various baselines on both the REFinD and TACRED datasets. Overall, incorporating SSDP-SEM yields consistent improvements in accuracy, precision, recall, and F1 scores. Notably, models that use SSDP-SEM achieve larger gains on REFinD than on TACRED, suggesting that sentiment-aware strategies are particularly effective in financial texts, where sentiment plays a more prominent role.

\begin{table*}[ht]
\belowrulesep=0pt
\aboverulesep=0pt
\centering
\caption{Ablation studies of ASP task with variant ISL on models we reproduced. Evaluated on the REFinD and TACRED dataset. 
\textbf{EP}, \textbf{SP}, \textbf{IS} refer to entity token position, SDP token position, implicit Sentiment token position respectively.} \label{table3}
\vspace{5pt}
\vspace{-0.4cm}
\begin{tabular}{>{\raggedright\arraybackslash}p{0.9cm}  >{\centering\arraybackslash}p{0.5cm}  >{\centering\arraybackslash}p{0.5cm}  >{\centering\arraybackslash}p{0.5cm}  |>{\centering\arraybackslash}p{0.7cm}  >{\centering\arraybackslash}p{0.7cm}  >{\centering\arraybackslash}p{0.7cm}  | > {\centering\arraybackslash}p{0.7cm}  >{\centering\arraybackslash}p{0.7cm}  >{\centering\arraybackslash}p{0.7cm}  | >{\centering\arraybackslash}p{0.7cm}  >{\centering\arraybackslash}p{0.8cm}  >{\centering\arraybackslash}p{0.7cm}  | > {\centering\arraybackslash}p{0.7cm}  >{\centering\arraybackslash}p{0.7cm}  >{\centering\arraybackslash}p{0.7cm}}

\hline
\toprule
\multirow{2}{*}{Dataset} &\multicolumn{3}{c}{ISL} & \multicolumn{3}{c}{Att-Bi-LSTM} & \multicolumn{3}{c}{R-BERT} & \multicolumn{3}{c}{RE-Improved} & \multicolumn{3}{c}{Casual} \\
& +EP & +SP & +IS & Pre & Rec & F1 & Pre & Rec & F1 & Pre & Rec & F1 & Pre & Rec & F1\\
\midrule
\multirow{3}{*}[-0.3cm]{REFinD} &$\checkmark$ & & 
& 73.84\% \newline \vspace{-1.3em} \textcolor{red}{\tiny(-1.43\%)\strut}
& 72.30\% \newline \vspace{-1.3em} \textcolor{red}{\tiny(-0.98\%)\strut} 
& 73.06\% \newline \vspace{-1.3em} \textcolor{red}{\tiny(-1.20\%)\strut}
& 75.18\% \newline \vspace{-1.3em} \textcolor{red}{\tiny(-1.48\%)\strut} 
& 76.81\% \newline \vspace{-1.3em} \textcolor{blue}{\tiny(+2.17\%)\strut} 
& 75.99\% \newline \vspace{-1.3em} \textcolor{blue}{\tiny(+0.35\%)\strut}
& 77.81\% \newline \vspace{-1.3em} \textcolor{red}{\tiny(-0.06\%)\strut} 
& 72.04\% \newline \vspace{-1.3em} \textcolor{red}{\tiny(-0.25\%)\strut}
& 74.81\% \newline \vspace{-1.3em} \textcolor{red}{\tiny(-0.17\%)\strut}
& 79.40\% \newline \vspace{-1.3em} \textcolor{blue}{\tiny(+0.75\%)\strut} 
& 73.61\% \newline \vspace{-1.3em} \textcolor{blue}{\tiny(+1.92\%)\strut}
& 76.40\% \newline \vspace{-1.3em} \textcolor{red}{\tiny(-0.66\%)\strut}
 \\[2.5ex]
&$\checkmark$ & $\checkmark$ & 
    & 72.09\% \newline \vspace{-1.3em} \textcolor{red}{\tiny(-3.18\%)\strut} 
    & 74.43\% \newline \vspace{-1.3em} \textcolor{blue}{\tiny(+1.15\%)\strut} 
    & 73.24\% \newline \vspace{-1.3em} \textcolor{red}{\tiny(-1.02\%)\strut} 
    & 75.62\% \newline \vspace{-1.3em} \textcolor{red}{\tiny(-1.04\%)\strut} 
    & 75.23\% \newline \vspace{-1.3em} \textcolor{blue}{\tiny(+0.59\%)\strut} 
    & 75.43\% \newline \vspace{-1.3em} \textcolor{red}{\tiny(-0.21\%)\strut}
    & 75.70\% \newline \vspace{-1.3em} \textcolor{red}{\tiny(-2.17\%)\strut} 
    & 74.77\% \newline \vspace{-1.3em} \textcolor{blue}{\tiny(+2.48\%)\strut}
    & 75.23\% \newline \vspace{-1.3em} \textcolor{blue}{\tiny(+0.25\%)\strut}
    & 78.86\% \newline \vspace{-1.3em} \textcolor{blue}{\tiny(+0.21\%)\strut} 
    & 74.89\% \newline \vspace{-1.3em} \textcolor{red}{\tiny(-0.64\%)\strut} 
    & 76.83\% \newline \vspace{-1.3em} \textcolor{red}{\tiny(-0.23\%)\strut}  
 \\[2.5ex]
&$\checkmark$ & $\checkmark$ & $\checkmark$ 
    & 75.27\% & 73.28\% & 74.26\% & 76.66\%  & 74.64\% & 75.64\% & 77.87\% & 72.29\% & 74.98\% & 78.65\%  & 75.53\% & 77.06\% 
 \\[1ex]
 
\midrule

\multirow{3}{*}[-0.3cm]{TACRED} &$\checkmark$ & & 
& 67.05\% \newline \vspace{-1.3em} \textcolor{blue}{\tiny(+1.38\%)\strut}
& 47.13\% \newline \vspace{-1.3em} \textcolor{red}{\tiny(-0.63\%)\strut} 
& 55.38\% \newline \vspace{-1.3em} \textcolor{red}{\tiny(-0.08\%)\strut}
& 69.32\% \newline \vspace{-1.3em} \textcolor{red}{\tiny(-2.28\%)\strut} 
& 67.55\% \newline \vspace{-1.3em} \textcolor{blue}{\tiny(+1.11\%)\strut} 
& 68.42\% \newline \vspace{-1.3em} \textcolor{red}{\tiny(-0.50\%)\strut}
& 67.43\% \newline \vspace{-1.3em} \textcolor{red}{\tiny(-1.40\%)\strut} 
& 68.93\% \newline \vspace{-1.3em} \textcolor{blue}{\tiny(+0.52\%)\strut}
& 68.17\% \newline \vspace{-1.3em} \textcolor{red}{\tiny(-0.49\%)\strut}
& 70.27\% \newline \vspace{-1.3em} \textcolor{red}{\tiny(-1.23\%)\strut} 
& 66.53\% \newline \vspace{-1.3em} \textcolor{red}{\tiny(-0.40\%)\strut}
& 68.35\% \newline \vspace{-1.3em} \textcolor{red}{\tiny(-0.84\%)\strut}
 \\[2.5ex]
&$\checkmark$ & $\checkmark$ & 
    & 58.98\% \newline \vspace{-1.3em} \textcolor{red}{\tiny(-6.69\%)\strut} 
    & 53.11\% \newline \vspace{-1.3em} \textcolor{blue}{\tiny(+5.35\%)\strut} 
    & 55.89\% \newline \vspace{-1.3em} \textcolor{blue}{\tiny(+0.59\%)\strut}  
    & 72.03\% \newline \vspace{-1.3em} \textcolor{blue}{\tiny(+0.43\%)\strut} 
    & 65.83\% \newline \vspace{-1.3em} \textcolor{red}{\tiny(-0.61\%)\strut} 
    & 68.79\% \newline \vspace{-1.3em} \textcolor{red}{\tiny(-0.13\%)\strut}
    & 69.00\% \newline \vspace{-1.3em} \textcolor{red}{\tiny(-0.17\%)\strut} 
    & 68.05\% \newline \vspace{-1.3em} \textcolor{red}{\tiny(-0.36\%)\strut}
    & 68.52\% \newline \vspace{-1.3em} \textcolor{red}{\tiny(-0.14\%)\strut}
    & 69.11\% \newline \vspace{-1.3em} \textcolor{red}{\tiny(-2.39\%)\strut} 
    & 68.78\% \newline \vspace{-1.3em} \textcolor{blue}{\tiny(+1.73\%)\strut} 
    & 68.95\% \newline \vspace{-1.3em} \textcolor{red}{\tiny(-0.24\%)\strut}  
 \\[2.5ex]
&$\checkmark$ & $\checkmark$ & $\checkmark$ 
    & 65.67\% & 47.76\% & 55.30\%   & 71.60\%  & 66.44\% & 68.92\% & 68.83\%  & 68.41\% & 68.66\%  & 71.50\%  & 67.05\% & 69.19\% 
 \\[1ex]
\hline
\toprule

\end{tabular}
\vspace{-0.4cm}
\end{table*}

On the REFinD dataset, integrating SSDP-SEM led to notable improvements for all models. For instance, Att-Bi-LSTM, an earlier soft attention model, benefited from SSDP-SEM, with its F1 score increasing from 72.93\% to 74.26\%. This result is noteworthy, given that Att-Bi-LSTM does not rely on multi-head attention yet still leverages our method effectively. Other reproduced models, such as R-BERT, RE-Improved, and Casual, also witnessed considerable gains in most metrics when augmented with SSDP-SEM. In particular, Casual with our module achieved an F1 score of 77.06\%, closely approaching the top-performing TrNP \cite{li2024enhancing}, and surpassed specialized BERT variants including FinBERT \cite{liu2021finbert} and SpanBERT \cite{joshi2020spanbert}.

On the TACRED dataset, the influence of ISL varied across models. While some approaches (e.g., Att-Bi-LSTM and RE-Improved) experienced minor performance declines in certain indicators under specific ISL configurations, others (notably R-BERT and Casual) gained substantial benefits from SSDP-SEM. Casual showed a notable increase in F1 score from 68.17\% to 69.19\%, demonstrating that a well-tuned ISL approach can enhance RE even in broad-domain contexts. At 69.19\%, Casual with SSDP-SEM outperformed several position-aware (PA-LSTM\cite{zhang2017position}) and graph-based (GCN\cite{guo2019attention}, CGCN\cite{zhang2018graph}) models, as well as noise-robust baselines like PURE\cite{zhong2020frustratingly} and Clean-LaVe\cite{wang2024use}. However, the model did not surpass the leading specialized approaches in highly noisy or complex scenarios. These findings imply that further integration of advanced features or specialized data augmentation techniques could boost the robustness and adaptability of RE models.

\section{Analyses and Discussion}

\subsection{Analysis of Sentiment Sensitivity}

\begin{figure}[H]
\centering
\vspace{-0.2cm}
\includegraphics[scale=0.3]{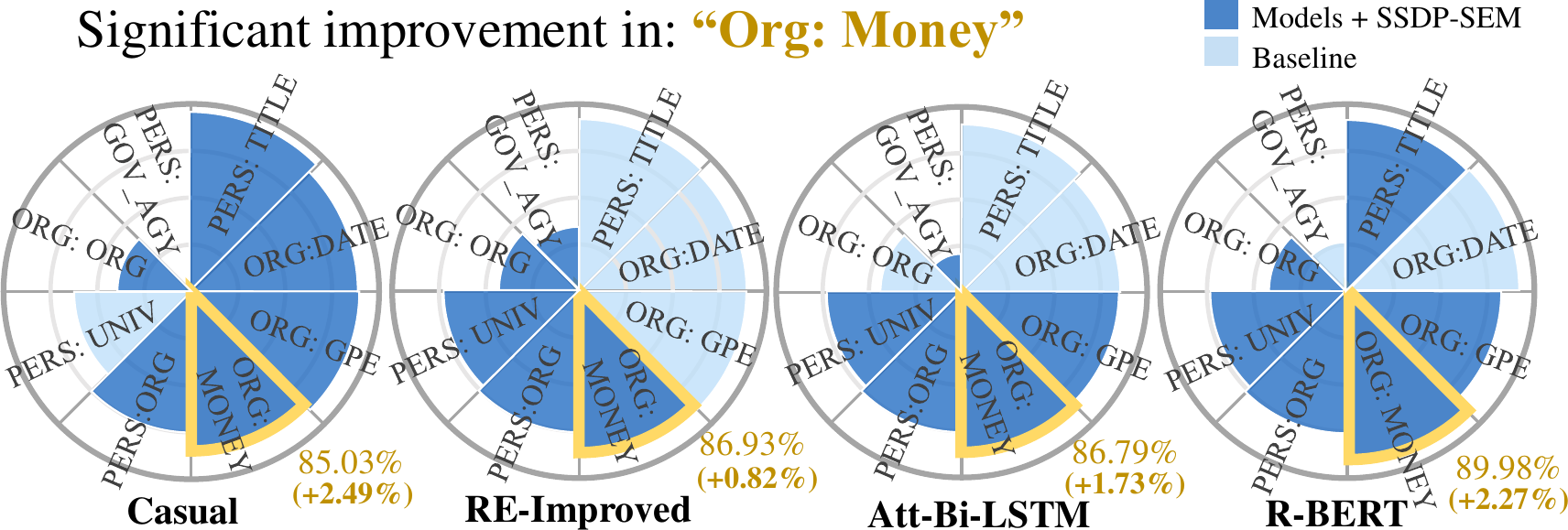}
\vspace{-0.1cm}
\caption{F1 score distribution by entity pairs on REFinD, indicating the performance of the baseline models and the models after integrating SSDP-SEM.}
\vspace{-0.3cm}
\label{figure5}
\end{figure}

To analyze the sentiment sensitivity of financial instance, we follow the previous work \cite{vardhan2023imetre}, which utilizes the entity pair classification strategy, to analyze the financial related entity-pair performance, as shown in Figure \ref{figure5}.

Four reproduced models with SSDP-SEM outperform all baselines, especially for sentiment-sensitive entity pairs (e.g., ORG:ORG and ORG:MONEY). Relations included in these entity pairs, such as ``revenue\_of", ``profit\_of", and ``agreement\_with", often reflect changes in financial performance or organizational status, both of which are inherently tied to sentiment---positive or negative reactions can drastically alter how these relationships are understood.

As visualized in the first circle of Figure \ref{figure5}, our method achieves an F1 score of 85.03\% on ``ORG: MONEY" on Casual, surpassing the baseline by +2.49\%. This phenomenon is reflected in the other three circles. This improvement demonstrates the framework's ability to leverage sentiment signals to disambiguate and correctly classify sentiment-sensitive entity pairs.

\subsection{Ablation Study of ASP Using ISL}

This section presents an ablation study examining the ASP task using ISL with three key semantics: EP, SP and IS token positions. Table \ref{table3} reports results on both REFinD and TACRED, illustrating how each position influences the model performance. Overall, while each individual component contributes to performance gains, removing the sentiment factor (\textit{i.e.}, IS) results in the most substantial drop, highlighting sentiment’s pivotal role in financial RE. Below, we discuss two other specific settings that illustrate how omitting certain components affects the performance:
\begin{itemize}
    
    \item \textbf{ISL removing IS (SPL):} When IS is removed, SPL focuses solely on SDP without sentiment guidance. Although SDP filtering helps reduce noise—particularly in financial texts—omitting sentiment leads to a marked performance decrease on both REFinD and TACRED compared to configurations that include IS. This suggests that, while syntactic cues remain valuable, sentiment information is critical for capturing the nuanced relationships common in financial discourse.

    \item \textbf{ISL removing EP and IS (EPL):} When EP and IS are excluded, EPL directs the model to consider only entity token positions, limiting its focus to entity dependencies. Owing to its reduced supervisory scope, this configuration substantially lowers overall performance. For example, as shown in Table \ref{table3}, Att-Bi-LSTM’s F1 score on REFinD decreases by 1.28\% when only EPL is used, highlighting the drawbacks of removing both sentiment and syntactic cues.

\end{itemize}

\subsection{Effectiveness of SAIB}
We apply SAIB regularization to encourage baseline models to focus on sentiment-related features by imposing sparse attention weights. As described in Formula \ref{eq_ib}, we introduce a sentiment token to compute attention weights, aiming to enhance the model focus on sentiment-sensitive instances. 

\begin{figure}[H]
\centering
\vspace{-0.1cm}
\includegraphics[scale=0.33]{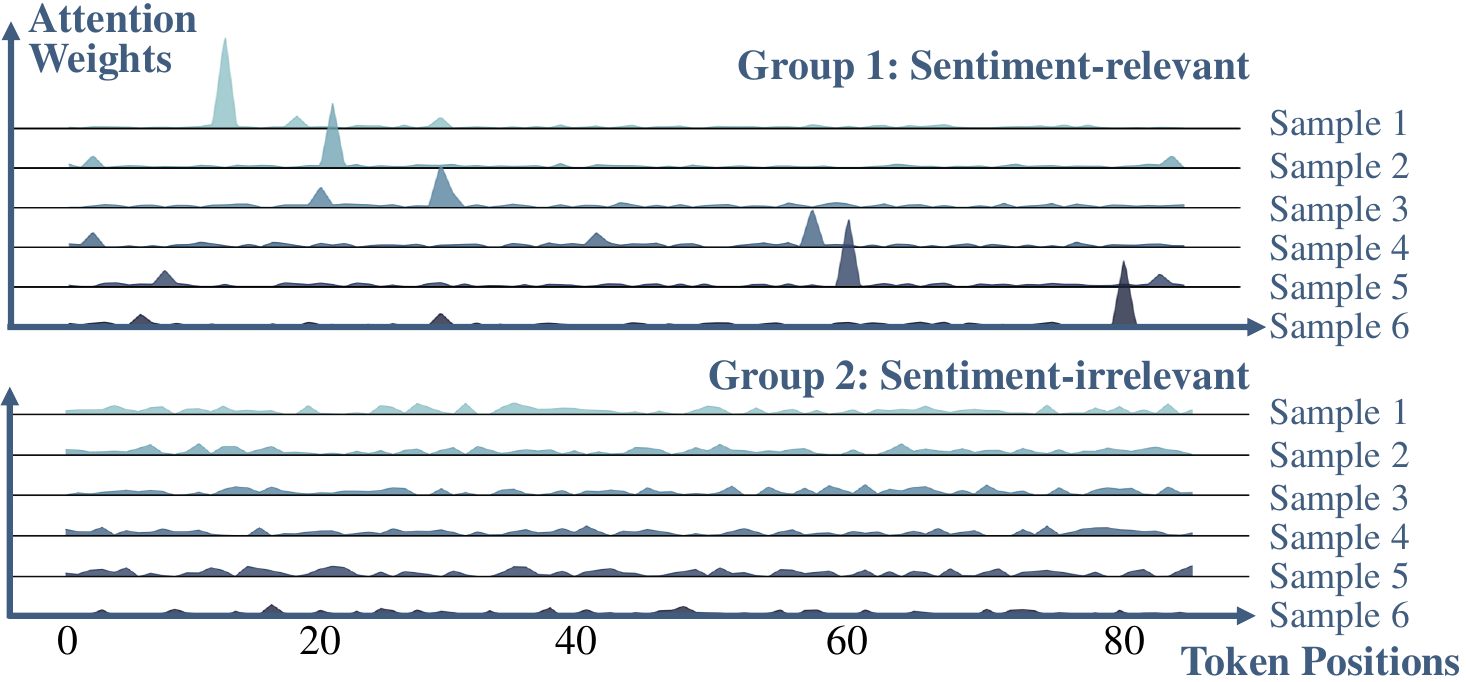}
\vspace{-0.3cm}
\caption{Visualization of SAIB attention weights. Group 1 represents attention weights for ``ORG:MONEY" entity pairs, while Group 2 includes other relation instances. }
\vspace{-0.2cm}
\label{fig_saib}
\end{figure}

To evaluate this effect, we visualize the attention weights on different instances, particularly “ORG:MONEY” entity pair and other cases, as shown in Figure \ref{fig_saib}. The results indicate that sentiment-related instances exhibit sparser attention distributions, whereas non-sentiment instances do not. This confirms that SAIB effectively guides the model to emphasize sentiment-driven entity relationships, thereby achieving our intended goal.

\subsection{Case Analysis}
\begin{figure}[h]
\centering
\vspace{-0.1cm}
\includegraphics[scale=0.16]{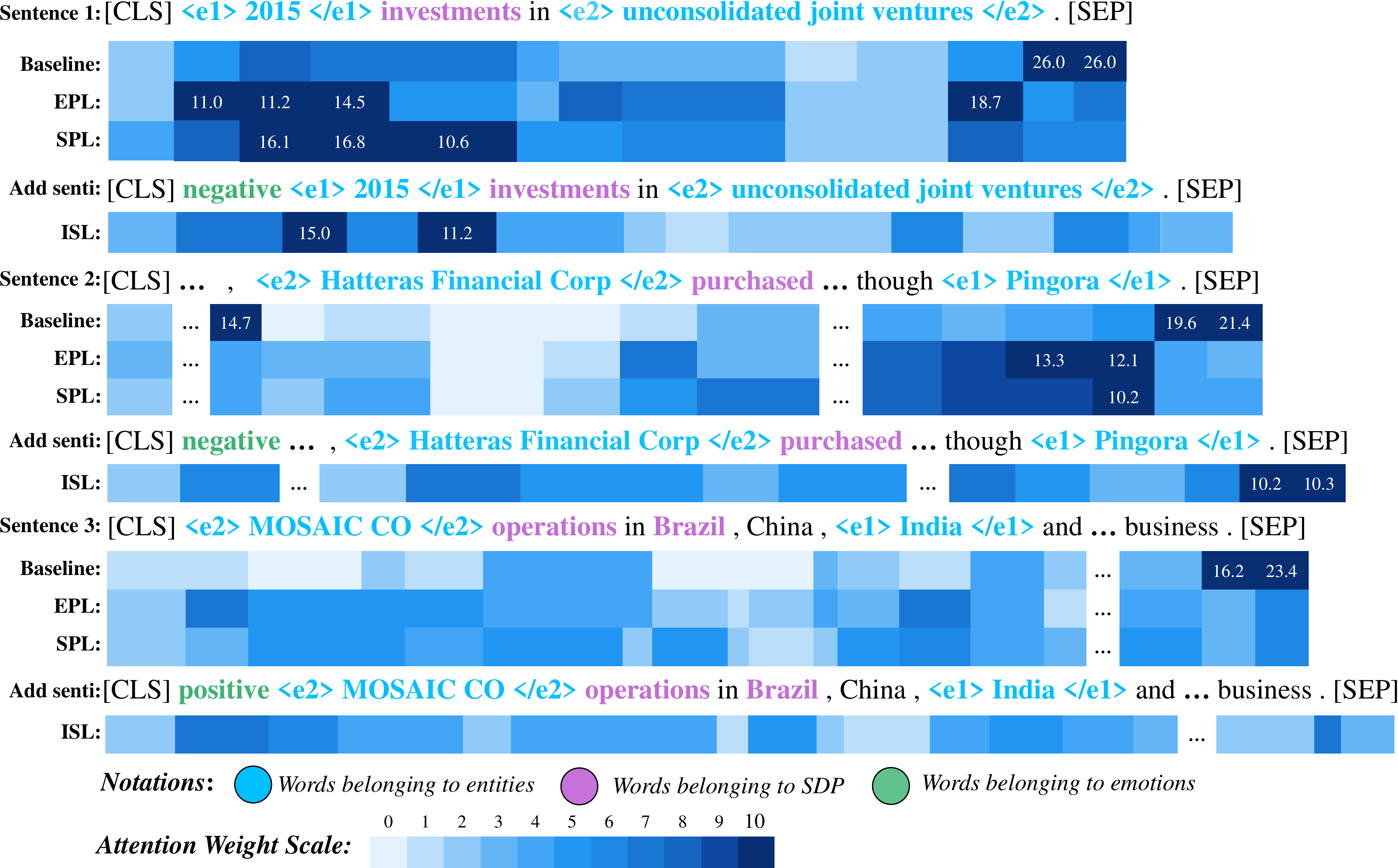}
\vspace{-0.6cm}
\caption{Attention distribution visualization across baseline, ISL and their corresponding EPL and SPL. Darker shades indicate higher attention levels, focusing on entity tokens, SDP tokens, and sentiment tokens.}
\vspace{-0.5cm}
\label{figure4}
\end{figure}

In this section, we analyze three specific sentences to demonstrate how multi-head attention allocation changes from the baseline to the ASP task with various signal configurations (EPL, SPL, and ISL) and how this enhances the model's performance. Figure \ref{figure4} visualizes the attention distribution from Casual.


In the third sentence of Figure \ref{figure4}, the baseline model distributes attention broadly across the sentence, without a clear focus on key components. For example, the first-line visualization of the third sentence fails to prioritize core entities like ``MOSAIC CO" and ``India", resulting in vague semantic understanding and inefficiency.

With EPL, which adds entity token positions, the model starts to focus more on subjective and objective entities. This shift improves the model’s ability to concentrate on entity-centric contexts, suppressing irrelevant tokens such as "in" and "are", although it still lacks the refinement needed to capture the full relations between entities.

As SPL is applied, the model improves its understanding of the syntactic structure, with greater attention placed on “operations” as the relational term linking “MOSAIC CO” to geographic entities. By aligning attention with this syntactic structure, the structural relationship between the company and its operational locations is better captured, leading to a more precise RE between the subject, verb, and object.

Finally, with ISL, sentiment signals are introduced into the attention distribution. ISL shifts the attention weights towards the words ``positive" and ``operations", effectively capturing the nature of the business performance---positive, and indicating a favorable view of MOSAIC CO's operations. As the positive sentiment guides the model’s focus on the fundamental business operations between MOSAIC CO and its operational locations, the model can extract the relation---“ORG:GPE:operations\_in” more accurately.



\subsection{Error Analysis}
While our method generally enhances performance, its benefits are not uniform across all financial instances. As shown in Figure \ref{figure5}, SSDP-SEM significantly improves RE for “ORG:MONEY” instances but shows no improvement for “ORG:DATA” cases across all baselines. This discrepancy suggests that while sentiment-aware features are useful for monetary relationships, they may be less effective for other entity pairs.

Even within “ORG:MONEY” instances, some cases remain challenging. One key issue arises when conflicting sentiment cues appear in the same sentence, making it difficult for the model to focus on the sentiment most relevant to the target relation. The following false negative example illustrates this issue:
\begin{itemize}
    \item \textbf{False Negative Example: }\textit{Despite facing significant losses, mCig, Inc. saw a substantial increase in revenue, reporting an additional \$1.3 billion from its core operations. } (\textbf{Positive Sentiment})
    
\end{itemize}

The correct relation should be “ORG:MONEY:revenue\_of”, linking “mCig, Inc.” (ORG) and “\$1.3 billion” (MONEY). However, all models incorporating SSDP-SEM incorrectly predicted “no\_relation”, failing to identify the positive sentiment implied by “substantial increase in revenue”. The presence of conflicting sentiments—negative (from “facing significant losses”) and positive (from “substantial increase in revenue”)—likely misled the model by obscuring the most relevant syntactic and emotional cues. This outcome highlights a broader limitation of SSDP-SEM: its reliance on sentiment signals can lead to misclassification when contradictory sentiments coexist.



\section{Conclusion}

We present SSDP-SEM, a pluggable framework that enhances financial RE by integrating sentiment analysis and SDP information. By incorporating sentiment-aware supervisory signals through the ASP task, SSDP-SEM refines attention mechanisms to capture both syntactic and emotional cues in financial texts. Our experiments demonstrate significant performance gains on the sentiment-sensitive REFinD dataset while maintaining strong generalization on TACRED, highlighting SSDP-SEM's adaptability across financial-specific and general-purpose RE tasks. Additionally, our analyses reveal the impact of different ASP configurations with SAIB regularization, emphasizing SSDP-SEM's effectiveness in handling sentiment-sensitive entity relationships, and offering insights for future sentiment analysis in financial contexts.





\bibliographystyle{IEEEtran}
\small\bibliography{reference}

\end{document}